\PassOptionsToPackage{silent}{fontspec}
\documentclass[cameraready]{template/mac_automl_xmu_blue}

\graphicspath{{./}}

\usepackage[toc,page,header]{appendix}

\usepackage{tabularx}
\newcommand{\myparagraph}[1]{\par\smallskip\noindent\textbf{#1}\hspace{1.2ex}}

\newcommand{\suppanchor}[1]{\hypertarget{#1}{}}
\newcommand{\supplink}[2]{\hyperlink{#2}{#1}}

\tcbset{
  aibox/.style={
    width=\linewidth,
    top=5pt,
    bottom=5pt,
    colback=blue!4!white,
    colframe=black,
    colbacktitle=black,
    enhanced,
    center,
    attach boxed title to top left={yshift=-0.08in,xshift=0.12in},
    boxed title style={boxrule=0pt,colframe=white},
    breakable,
  }
}
\renewtcolorbox{AIbox}[2][]{aibox,title=#2,#1}

\setleftheadercontent{%
  \headerlogospace{1.6mm}%
  \headerlogo[-0.08\height]{15.2mm}{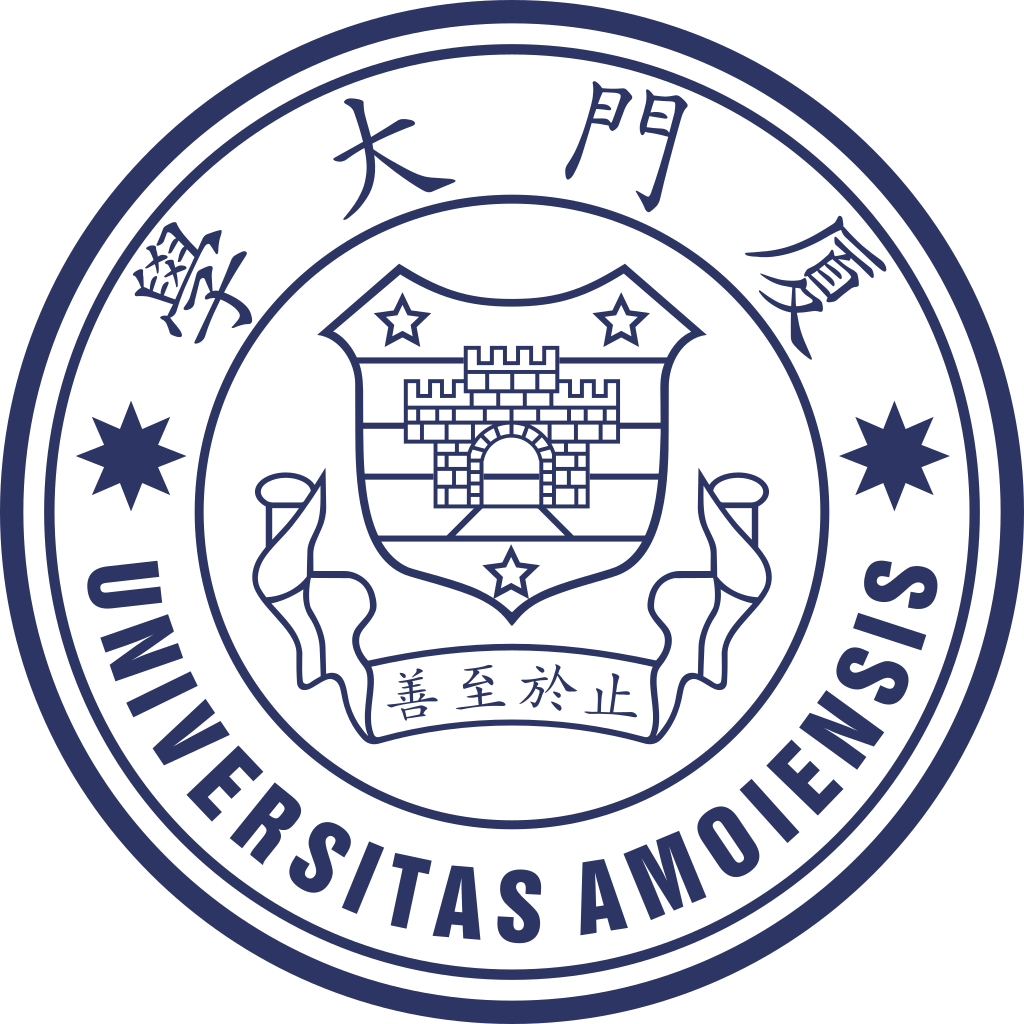}%
  \headerlogospace{2.8mm}%
  \headerlogo[-0.06\height]{12.6mm}{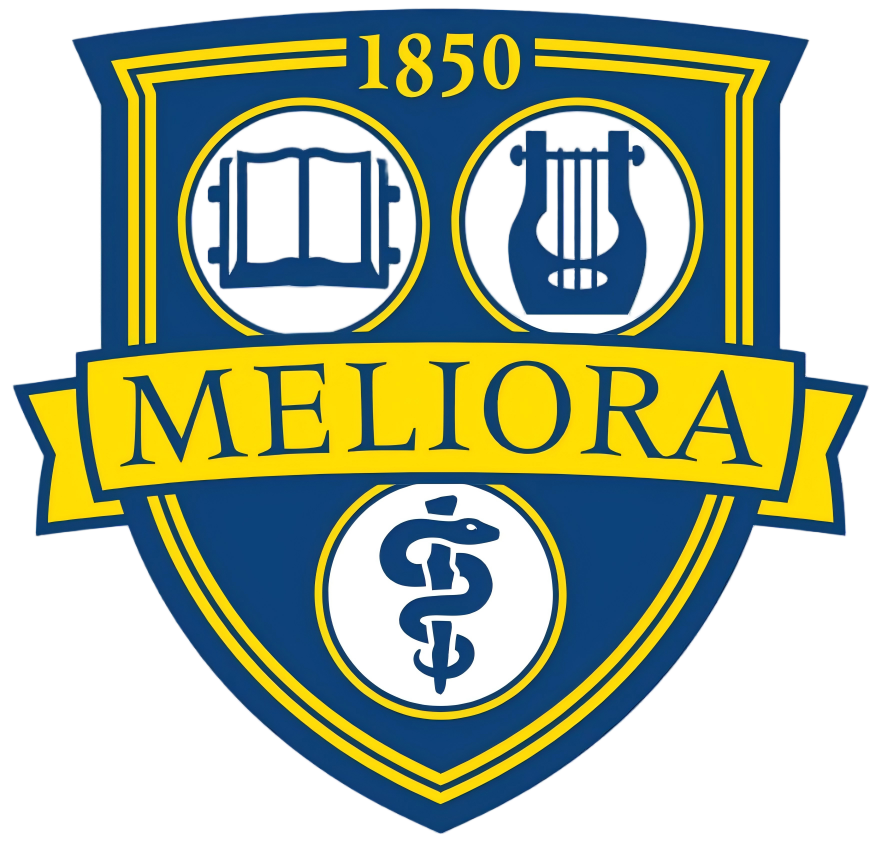}%
  \headerlogospace{2.8mm}%
  \headerlogo[-0.05\height]{12.8mm}{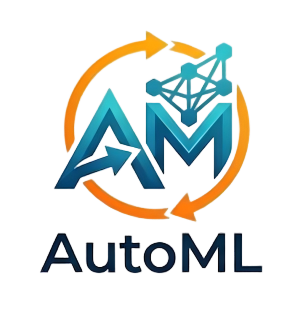}%
}
\setrightheadericon{}
\setrunningheadericon{%
  \headerlogospace{1.2mm}%
  \headerlogo[-0.03\height]{8.2mm}{assets/branding/automl.png}%
}
\setheadergroupname{MAC-AutoML}

\title{SocialOmni: Benchmarking Audio-Visual\\Social Interactivity in Omni Models}

\setfrontauthors{%
  \authorrow{\authorentry{Tianyu Xie}{1},\authorsep \authorentry{Jinfa Huang}{2},\authorsep \authorentry{Yuexiao Ma}{3},\authorsep \authorentry{Rongfang Luo}{4},\authorsep \authorentry{Yan Yang}{4}}%
  \authorrow{\authorentry{Qingchuan Ma}{1},\authorsep \authorentry{Wang Chen}{1},\authorsep \authorentry{Yuhui Zeng}{1},\authorsep \authorentry{Yixuan Zou}{1},\authorsep \authorentry{Zhiqiang Lu}{1}}%
  \authorrow{\authorentry{Ruize Fang}{1},\authorsep \authorentry{Jiebo Luo}{2},\authorsep \authorentry{Rongrong Ji}{1},\authorsep \authorentry{Xiawu Zheng}{1,\corremailmark}}%
}
\setfrontaffiliations{%
  \affiliationline{\authormark{1}Xiamen University}
  \affiliationline{\authormark{2}University of Rochester}
  \affiliationline{\authormark{3}Nanyang Technological University}
  \affiliationline{\authormark{4}Sichuan Agricultural University}
}
\setfrontcontact{%
  \contactline{\textbf{\corremailmark\ Corresponding Author}}
}
\usecustomauthorlayout

\checkdata[Email]{Xiawu Zheng \href{mailto:zhengxiawu@xmu.edu.cn}{\nolinkurl{zhengxiawu@xmu.edu.cn}}}
\checkdata[Code]{\href{https://github.com/MAC-AutoML/SocialOmni}{\nolinkurl{github.com/MAC-AutoML/SocialOmni}}}
\checkdata[Dataset]{\href{https://huggingface.co/datasets/alexisty/SocialOmni}{\nolinkurl{huggingface.co/datasets/alexisty/SocialOmni}}}
\hypersetup{pdftitle={SocialOmni: Benchmarking Audio-Visual Social Interactivity in Omni Models},pdfauthor={Tianyu Xie, Jinfa Huang, Yuexiao Ma, Rongfang Luo, Yan Yang, Qingchuan Ma, Wang Chen, Yuhui Zeng, Yixuan Zou, Zhiqiang Lu, Ruize Fang, Jiebo Luo, Rongrong Ji, Xiawu Zheng}}

\abstract{Evaluating omni-modal large language models (OLMs) in multi-party dialogue requires more than answer correctness on pre-segmented inputs. We introduce \textbf{SocialOmni}, an offline diagnostic benchmark that separates three turn-level decisions: identifying \textit{who} is speaking, deciding \textit{when} a designated participant should enter at an annotated query time, and determining \textit{how} that participant should continue the dialogue. SocialOmni contains 2,000 perception items and a quality-controlled core split of 200 interaction-generation items, including naturally occurring speaker-visibility mismatches. Each item is independently checked by three human annotators. Evaluated systems receive only query-time-bounded multimodal evidence, while manually verified reference continuations are reserved for response judging. A complete three-judge ensemble scores every eligible response, with leave-one-judge-out and family-sensitivity audits. Across 11 OLMs, rankings vary substantially by axis, and coverage-adjusted scores reveal when high conditional response quality depends on selective turn entry. The protocol does not measure persistent streaming state or wall-clock latency.
}

\begin{document}

\maketitle

\section{Introduction}
\label{sec:intro}

Current OLM evaluation under-specifies the integrated decisions required for social dialogue. Omni-modal large language models combine audio, vision, and text within a unified generation interface~\citep{hurst2024gpt,comanici2025gemini,Qwen3-Omni,fu2025vita,li2025baichuan,ye2025omnivinci}, but dialogue success also requires recognizing dynamic cues, selecting an appropriate entry point, and producing a socially coherent response. As summarized in~\Cref{tab:omni_benchmarks}, most benchmarks remain centered on static, accuracy-oriented understanding tasks~\citep{li2024omnibench,zhou2025worldsense,yang2025omnieval,li2025omnivideobench}, leaving the integrated interaction decision largely unmeasured.

The missing unit of analysis is the integrated conversational turn. At a single dialogue state, a model must identify the active speaker, determine whether a designated participant should enter, and produce a grounded continuation. SocialOmni targets this unit directly rather than introducing another static comprehension leaderboard; the Related Work section contrasts it with prior benchmark families and motivates our interaction-centered protocol.

%---------------------------------------------------------------%

This missing turn-level evaluation matters because semantic correctness does not guarantee successful interaction. A delayed entry, premature interruption, or incoherent continuation can degrade user experience even when the propositional content is accurate~\citep{sacks1974turntaking,skantze2021turntaking}. Correctness-only evaluation can therefore over-reward offline comprehension while under-penalizing interaction failures~\citep{salmonn2024omni}. SocialOmni addresses this gap through three coordinated diagnostics:

\begin{AIbox}{\textbf{SocialOmni: Dimensions of Social Interactivity}}
\textbf{Who}: identify the active speaker in a marked interval.\par
\textbf{When}: decide whether the target should speak now.\par
\textbf{How}: generate an appropriate reply for the target role.
\end{AIbox}

%---------------------------------------------------------------%

SocialOmni organizes the integrated turn as a diagnostic sequence from audio-visual grounding to turn-entry decision and contextual continuation. \emph{Who} requires fine-grained speaker--content alignment, \emph{when} integrates prosodic, lexical, and visual cues at an annotated state, and \emph{how} tests whether the resulting continuation remains role-consistent and contextually appropriate. The benchmark contains 2,000 perception items and a quality-controlled core set of 200 interaction-generation items across 15 dialogue categories in four broad domains, including naturally occurring speaker-visibility mismatches. It does not reproduce a persistent full-duplex session or measure system latency.

%---------------------------------------------------------------%

Evaluation across 11 OLMs supports a diagnostic profile rather than a single competence score. Model leadership changes across speaker attribution, turn-entry classification, gold-state generation, and coverage-adjusted response quality, so gains on one measure do not establish robustness on others. These descriptive rank reversals show that an understanding-centric score alone cannot characterize interaction behavior across these coupled decisions.

%---------------------------------------------------------------%
Our contributions are threefold:

\begin{enumerate}[leftmargin=*,nosep,label=\roman*)]
    \item \textbf{Diagnostic Benchmark.} We evaluate 2,200 items across the \emph{who}, \emph{when}, and \emph{how} diagnostics.
    \item \textbf{Auditable Response Evaluation.} Three judges score every response, with judge/family sensitivity audits.
    \item \textbf{Grounding Analysis.} Visibility strata diagnose attribution failures without causal claims.
\end{enumerate}

\begin{table*}[t]
\caption{Positioning of OLM benchmarks by diagnostic task, speaker-visibility stratification, and temporal granularity. \cmark~denotes an explicit task, \pmark~an indirect proxy, and \xmark~no explicit task.}
\label{tab:omni_benchmarks}
  \centering
  {\small
  \setlength{\tabcolsep}{2pt}
  \renewcommand{\arraystretch}{1.12}
  \newcolumntype{L}[1]{>{\raggedright\arraybackslash}m{#1}}
  \newcolumntype{C}[1]{>{\centering\arraybackslash}m{#1}}
  \begin{tabular}{@{}L{0.33\textwidth} C{0.17\textwidth} C{0.06\textwidth} C{0.065\textwidth} C{0.055\textwidth} C{0.11\textwidth} C{0.14\textwidth}@{}}
    \toprule
    \textbf{Benchmark} & \textbf{Types} & \textbf{Who} & \textbf{When} & \textbf{How} & \makecell{\textbf{Visibility} \\ \textbf{strata}} & \makecell{\textbf{Temporal} \\ \textbf{granularity}} \\
    \midrule
    OmniBench~\citep{li2024omnibench}               & Understanding & \xmark & \xmark & \xmark & \xmark & \fb\eb\eb\eb \\
    OmniVideoBench~\citep{li2025omnivideobench}      & Understanding & \pmark & \pmark & \xmark & \xmark & \fb\fb\eb\eb \\
    WorldSense~\citep{zhou2025worldsense}            & Understanding & \pmark & \pmark & \xmark & \xmark & \fb\fb\eb\eb \\
    OmniEval~\citep{yang2025omnieval}                & Understanding & \pmark & \pmark & \xmark & \xmark & \fb\eb\eb\eb \\
    Daily-Omni~\citep{zhou2025dailyomni}             & Understanding & \pmark & \pmark & \xmark & \xmark & \fb\fb\eb\eb \\
    JointAVBench~\citep{chao2025jointavbench}        & Understanding & \pmark & \pmark & \xmark & \xmark & \fb\fb\eb\eb \\
    OmniMMI~\citep{wang2025omnimmi}                  & Interaction   & \pmark & \pmark & \pmark & \xmark & \fb\fb\fb\eb \\
    Omni-SafetyBench~\citep{pan2025omnisafetybench}  & Understanding & \xmark & \xmark & \xmark & \xmark & \fb\eb\eb\eb \\
    \rowcolor{omniaccent!18}
    {\color{omniaccent!90!black}\textbf{SocialOmni (Ours)}} &
    {\color{omniaccent!90!black}\textbf{Interaction}} &
    \cmark &
    \cmark &
    \cmark &
    \cmark &
    {\color{omniaccent!90!black}\textbf{\fb\fb\fb\eb}} \\
    \bottomrule
  \end{tabular}}

\end{table*}

%---------------------------------------------------------------%

\section{Related Work}
\label{sec:related}
\myparagraph{Omni-Modal Large Language Models.}
OLM development has broadened multimodal generation faster than interaction-focused evaluation. The field has progressed from perception-centric paradigms such as CLIP~\citep{radford2021clip} and instruction-tuned VLMs such as Flamingo and LLaVA~\citep{alayrac2022flamingo,liu2023llava,liu2023llava15} to systems that couple text, vision, and audio~\citep{hurst2024gpt,comanici2025gemini,Qwen3-Omni,fu2025vita,ye2025omnivinci,li2025baichuan,xie2024mini}. Architectures now range from modular dispatch to native cross-modal generation~\citep{google2026geminiapi,xu2025qwen25omni,zhang2025mixture,liu2025comprehensive,salmonn2024omni}, yet evaluation remains dominated by isolated request--response trials. Speaker grounding, turn entry, and socially appropriate continuation therefore remain comparatively under-tested.

\myparagraph{Answer-Centric Benchmarks for OLMs.}
Answer-centric benchmarks establish perception and reasoning quality but do not operationalize conversational entry. Broad-coverage omni suites~\citep{li2024omnibench,li2025omnivideobench,zhou2025worldsense,yang2025omnieval} and modality-specific benchmarks~\citep{liu2023mmbench,yue2023mmmu,li2024seedbench,fu2025video,li2024mvbench,zhou2024mlvu,wang2025audiobench} evaluate question answering or retrieval over pre-segmented stimuli. Video benchmarks add temporal reasoning while retaining isolated prompt--response trials. Consequently, these suites do not test whether a designated participant should enter at a particular dialogue state or produce a fitting continuation.

\myparagraph{Behavior-Centric Benchmarks for OLMs.}\quad
Behavior-centric benchmarks move toward interaction, but typically isolate only part of the integrated turn. Social reasoning datasets~\citep{kong2025sivbench,mathur2025socialgenome} target multi-speaker inference, while spoken-dialogue and full-duplex benchmarks~\citep{arora2025talkingturns,lin2025fullduplexbench_turntaking,selvakumar2025multivox,wang2025voiceassistanteval,lin2026wearvox,jiang2025specific} emphasize turn-taking or interruption detection, often from audio alone. Multimodal interaction benchmarks~\citep{chowdhury2025amuse,zhou2025dailyomni,chao2025jointavbench,wang2025omnimmi,pan2025omnisafetybench} broaden audio-visual coverage. SocialOmni complements these efforts with one offline protocol spanning speaker attribution, fixed-time turn-entry classification, and conditional response generation. Its purpose is controlled cross-model diagnosis under a shared task definition and evaluation contract, not persistent-state or wall-clock latency evaluation.
%---------------------------------------------------------------%

\section{SocialOmni}
\label{sec:method}

SocialOmni evaluates an integrated conversational turn through three offline diagnostics. The tasks identify the active speaker (\emph{who}), decide whether a designated participant should enter at an annotated query time (\emph{when}), and generate an appropriate continuation after a correctly detected opportunity (\emph{how}). Every model receives query-time-bounded evidence under a common task definition, with a disclosed adapter when an API cannot accept joint audio and visual input. The protocol does not measure persistent streaming state, wall-clock latency, or deployment efficiency.

\begin{figure*}[t]
\centering
\includegraphics[width=\textwidth]{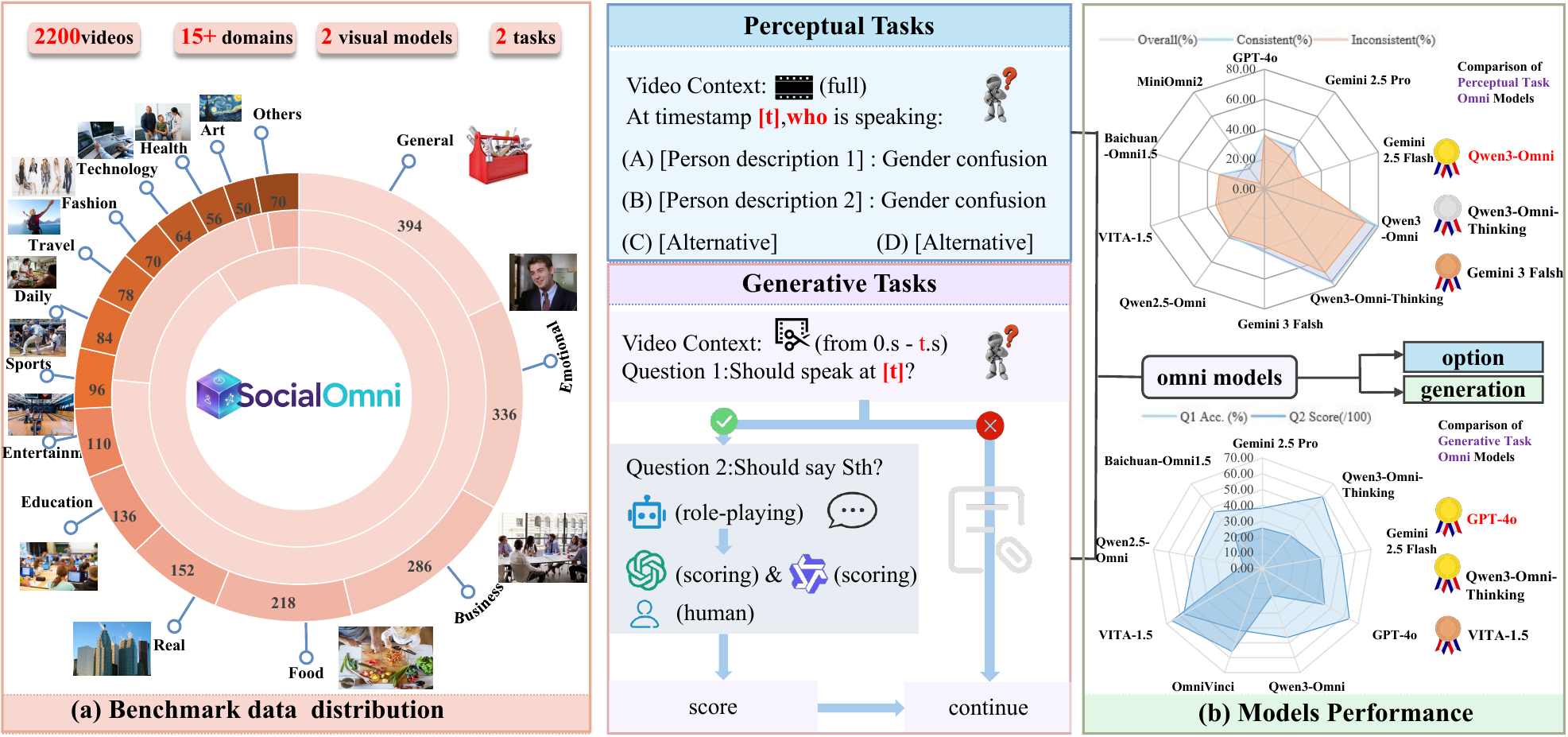}
\caption{SocialOmni overview. The benchmark uses two disjoint task splits totaling 2,200 items. The \emph{when} task is a fixed-time classification over an observed prefix, while \emph{how} evaluates responses only after a correctly detected gold-positive opportunity. The manually verified reference continuation is reserved for response judging. The profiles at right are illustrative; all quantitative rankings and values are reported in Table~\ref{tab:main_triplet}.}
\label{fig:socialomni_overview}
\end{figure*}

\subsection{Benchmark Construction}
\label{sec:dataset_construction}

Benchmark construction prioritizes dialogue clarity and reproducible item definitions. We compile search terms across 15 dialogue categories in four broad domains: entertainment, professional, daily life, and narrative. From more than 3,000 reviewed source videos, we retain 10--30\,s segments with clear audio, visible conversational context, and licenses compatible with reconstruction. A \emph{source video} may yield a clip, whereas a \emph{benchmark item} denotes one evaluated query constructed from a retained clip. The resulting benchmark contains 2,000 perception items and 200 interaction-generation items, with a mean clip duration of 25.0\,s.

Three-stage human verification defines every benchmark label. Each item is independently checked by three annotators drawn from a trained pool, covering speaker identity, spoken-content options, visibility, target-participant descriptions, query times, YES/NO turn-entry labels, and reference continuations. Cross-review and senior adjudication then resolve disagreements and verify the textual artifacts. The supplementary material reports annotation roles and agreement.

\subsection{Data Stratification}
\label{sec:benchmark_statistics}

The perception split stratifies speaker attribution by naturally occurring visibility conditions. It contains 1,725 \emph{speaker-visible} items, where the active speaker is visually attributable at query time, and 275 \emph{speaker-visibility-mismatch} items, including reaction shots, off-screen speech, partial occlusion, and weak identity cues. These cases are neither artificial audio swaps nor matched counterfactual interventions. The split therefore diagnoses grounding difficulty without supporting a causal estimate of controlled audio-visual conflict.

The interaction-generation split is a controlled aggregate diagnostic rather than a per-domain leaderboard. Each of its 200 items must provide an identifiable target participant, an unambiguous query state, sufficient multimodal context, and a human-written continuation. We consequently report aggregate results with uncertainty and treat language-, culture-, and domain-specific conclusions as out of scope.

\subsection{Task Design}
\label{sec:task_formulation}

\noindent\textbf{Who: speaker attribution.}\quad
The \emph{who} task separates speaker grounding from content selection. At query interval $t$, a model receives video $V$, audio $\mathcal{A}$, and four options formed by crossing speaker identity with spoken content. The options contain the correct identity--content pair and three distractors: wrong identity with correct content, correct identity with wrong content, and both wrong.

\noindent\textbf{When: fixed-time turn-entry classification.}\quad
The \emph{when} task evaluates turn entry at a single controlled dialogue state. For item $i$, annotators identify target participant $X_i$ with a visible description and select query time $t_i$. The model receives only prefix $(V_{i,\le t_i},\mathcal{A}_{i,\le t_i})$ and answers whether $X_i$ should begin a substantive turn using \texttt{A} (\texttt{YES}) or \texttt{B} (\texttt{NO}). Repeated prefix queries, first-response timing, and system latency remain outside the protocol.

\noindent\textbf{How: conditional response generation.}\quad
The \emph{how} task isolates continuation quality after a correctly detected opportunity. A model that correctly answers \texttt{YES} on a gold-positive item generates the next utterance for the same target participant $X_i$. Fillers and backchannels count as full turns only when the annotation targets that pragmatic function. The evaluated model never sees the human-authored reference continuation, which is reserved for response judging.

\begin{figure*}[t]
  \centering
  \includegraphics[width=\textwidth]{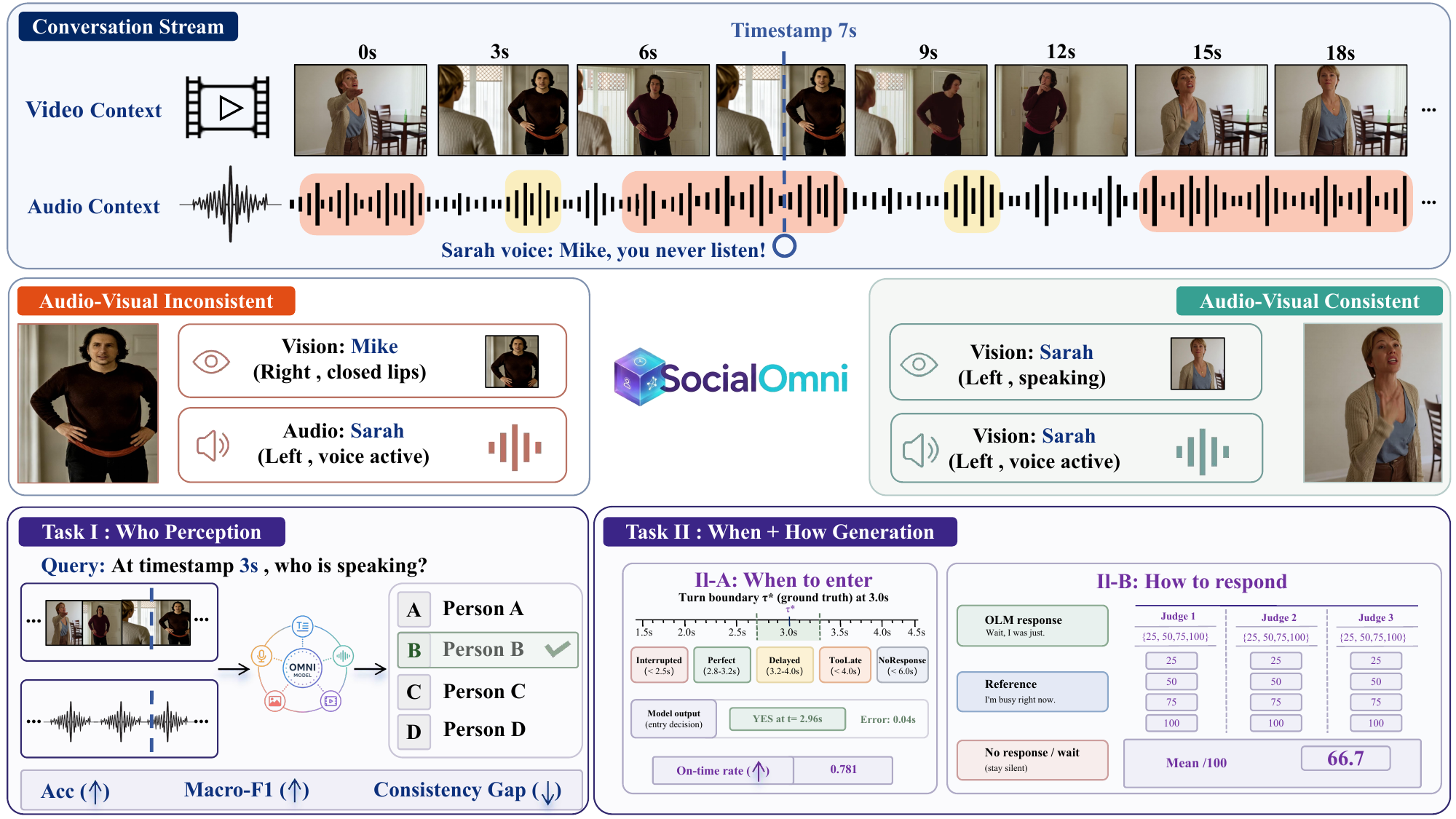}
  \caption{Study case for the three diagnostics. From one query-time-bounded conversation, SocialOmni tests speaker--content attribution (\emph{who}), whether the designated participant should enter at the annotated state (\emph{when}), and the role-consistent continuation produced after a correct positive decision (\emph{how}). ``Consistent'' and ``inconsistent'' denote observed speaker-visibility alignment, not synthetic audio swapping.}
  \label{fig:study_case}
\end{figure*}

\subsection{Evaluation Protocol}
\label{sec:eval_protocol}

Speaker-attribution metrics quantify both aggregate accuracy and visibility-conditioned difficulty. Non-parsable outputs count as incorrect; we report top-1 accuracy, macro-F1 over the four answer positions, and accuracy within the speaker-visible and visibility-mismatch strata. The gap $\Delta_{\rm vis}=\mathrm{Acc}_{\rm visible}-\mathrm{Acc}_{\rm mismatch}$ is descriptive; naturally occurring, unmatched strata cannot identify causal visibility effects across models and dialogue settings.

Turn-entry metrics characterize classification at the same annotated query states. We report accuracy, precision, recall, and macro-F1 over 200 binary decisions, with non-parsable outputs counted as incorrect. These metrics avoid imposing a universal timing window across dialogues.

Response-generation metrics separate conditional quality from opportunity coverage. Let $\mathcal{P}=\{i:y_i=\texttt{YES}\}$ denote gold-positive opportunities and $\mathcal{R}_m\subseteq\mathcal{P}$ the items on which model $m$ both predicts \texttt{YES} and returns a non-empty utterance. The three fixed judges independently score each response in $\{0,25,50,75,100\}$, and $q_{mi}$ is their complete mean. We define $\mathrm{Cov}^{+}_m=|\mathcal{R}_m|/|\mathcal{P}|$, $Q_{{\rm Ens},m}=\sum_{i\in\mathcal{R}_m}q_{mi}/|\mathcal{R}_m|$, and $Q_{{\rm joint},m}^{\rm Ens}=\mathrm{Cov}^{+}_mQ_{{\rm Ens},m}$. The conditional score describes produced responses, whereas $Q_{\rm joint}^{\rm Ens}$ assigns zero to missed positive opportunities. False-positive entry remains captured by \emph{when} precision, so $Q_{\rm joint}^{\rm Ens}$ is not a complete interaction utility for deployment.

Judge completeness and sensitivity checks make the response score auditable. Every eligible response requires all three judge scores; failed calls are retried rather than averaged over a smaller pool. Leave-one-judge-out and same-family removal test dependence on any single evaluator or model family in the ensemble under evaluation.

\begin{algorithm}[t]
\caption{Fixed-prefix \emph{when--how} evaluation}
\label{alg:when_how}
\begin{algorithmic}[1]
\REQUIRE Items $(V_i,\mathcal{A}_i,t_i,X_i,y_i)$; model $f$; fixed judge set $\mathcal{J}$
\FOR{each item $i$}
  \STATE Query $f$ on $(V_{i,\le t_i},\mathcal{A}_{i,\le t_i},X_i)$ for \texttt{YES}/\texttt{NO}
  \STATE Record the binary prediction; count non-parsable output as incorrect
  \IF{$y_i=\texttt{YES}$ and the prediction is \texttt{YES}}
    \STATE Generate target-participant response $\hat{s}_i$
    \STATE Score $\hat{s}_i$ with every judge in $\mathcal{J}$ on $\{0,25,50,75,100\}$
  \ENDIF
\ENDFOR
\STATE Report \emph{when} classification metrics and $\mathrm{Cov}^{+}$, $Q_{\rm Ens}$, $Q_{\rm joint}^{\rm Ens}$
\end{algorithmic}
\end{algorithm}

\section{Experiments}
\label{sec:experiments}

The experiments test both cross-model performance and the failure patterns hidden by aggregate scores. We ask how current OLMs compare across the three diagnostics and which conclusions change after coverage, modality, visibility, and judge sensitivity are exposed explicitly.

\IfFileExists{tables/extension_main_triplet.tex}{%
  \begin{table*}[t]
\caption{Primary strict-denominator results. $Q_{\rm Gold}$ forces generation at all 128 gold-positive opportunities; $Q_{\rm Ens}$ is conditional on a non-empty true-positive response; Cov.$^{+}$ and $Q_{\rm joint}^{\rm Ens}$ retain the model's entry policy.}
\label{tab:main_triplet}
\centering
\setlength{\tabcolsep}{2pt}
{\small
\begin{tabular*}{\textwidth}{@{\extracolsep{\fill}}llrrrrrr}
\toprule
Model & Interface & Who & When & $Q_{\rm Gold}$ & $Q_{\rm Ens}$ & Cov.$^{+}$ & $Q_{\rm joint}^{\rm Ens}$ \\\midrule
GPT-4o~\citep{hurst2024gpt} & Cascade & 35.05 & 50.50 & \textbf{77.15} & \textbf{76.50} & 30.47 & 23.31 \\
Gemini 2.5 Pro~\citep{comanici2025gemini} & Visual-only & 39.90 & 52.50 & 15.62 & 12.37 & 48.44 & 5.99 \\
Gemini 2.5 Flash~\citep{comanici2025gemini} & Visual-only & 33.70 & 55.50 & 21.35 & 21.36 & 42.97 & 9.18 \\
Gemini 3 Flash~\citep{google2025gemini3} & Visual-only & 48.10 & 49.50 & 13.93 & 17.61 & 34.38 & 6.05 \\
Gemini 3 Pro~\citep{google2025gemini3} & Visual-only & 45.40 & 52.00 & 21.55 & 25.00 & 32.03 & 8.01 \\
Qwen3-Omni~\citep{Qwen3-Omni} & Native AV & \textbf{74.65} & 58.00 & 44.47 & 42.59 & 63.28 & 26.95 \\
Qwen3-Omni-Thinking~\citep{Qwen3-Omni} & Native AV & 67.65 & 56.00 & 67.77 & 64.86 & 46.88 & 30.40 \\
Qwen2.5-Omni~\citep{xu2025qwen25omni} & Native AV & 40.95 & 60.50 & 36.00 & 33.14 & 67.97 & 22.53 \\
OmniVinci~\citep{ye2025omnivinci} & Native AV & 29.75 & 64.50 & 40.82 & 38.52 & 70.31 & 27.08 \\
VITA-1.5~\citep{fu2025vita} & Native AV & 32.05 & \textbf{65.50} & 48.37 & 49.93 & \textbf{88.28} & \textbf{44.08} \\
Baichuan-Omni-1.5~\citep{li2025baichuan} & Native AV & 22.45 & 36.50 & 18.36 & 19.27 & 12.50 & 2.41 \\
\bottomrule\end{tabular*}}

\end{table*}
}{%
  \input{tables/main_triplet}%
}

\subsection{Setup}
\label{sec:setup}

\noindent\textbf{Models.}\quad
The evaluation covers eleven commercial and open-source systems: GPT-4o~\citep{hurst2024gpt}; Gemini~2.5 Pro/Flash~\citep{comanici2025gemini} and Gemini~3 Flash/Pro Preview~\citep{google2025gemini3}; Qwen3-Omni, Qwen3-Omni-Thinking, and Qwen2.5-Omni~\citep{Qwen3-Omni,xu2025qwen25omni}; OmniVinci~\citep{ye2025omnivinci}; VITA-1.5~\citep{fu2025vita}; and Baichuan-Omni-1.5~\citep{li2025baichuan}. All use the common task contract through disclosed interfaces, with every attempted item retained in the strict denominator.

\noindent\textbf{Inputs.}\quad
Input handling preserves the query-time boundary while exposing interface differences. Native omni interfaces consume truncated audio-video media directly. GPT-4o uses a disclosed cascade: \texttt{gpt-4o-transcribe} processes the truncated audio, and the chat endpoint receives its text with sampled frames. The Gemini relay accepts sampled frames but not raw audio, so its rows are visual-only baselines. Released metadata records these differences for cross-model interpretation. Evaluated models never receive the manually verified context or reference continuation reserved for judges.

\noindent\textbf{Metrics.}\quad
The metrics preserve each diagnostic's interpretation. We use accuracy and macro-F1 for \emph{who}; accuracy, positive-class precision/recall/F1, and macro-F1 for \emph{when}; and $Q_{\rm Ens}$, Cov.$^{+}$, and $Q_{\rm joint}^{\rm Ens}$ for \emph{how}. Table~\ref{tab:main_triplet} reports point estimates; the supplementary material provides uncertainty and failure accounting for every evaluated system.

\subsection{Main Results}
\label{sec:main_results}

Performance is strongly axis dependent. Table~\ref{tab:main_triplet} assigns the best speaker attribution, turn-entry classification, conditional response quality, and coverage-adjusted response quality to different systems. These rank changes justify a diagnostic profile rather than a single social-competence score, but they do not establish causal independence among the axes.

Coverage adjustment materially changes the interpretation of response quality. The GPT-4o cascade obtains the highest conditional score ($76.50$) but responds to only $30.47\%$ of positive opportunities, reducing its joint score to $23.31$. Qwen3-Omni-Thinking reaches $64.86$ conditional and $30.40$ joint quality, whereas VITA-1.5 combines $49.93$ conditional quality with $88.28\%$ coverage to obtain the highest joint score ($44.08$). Conditional quality alone therefore overstates systems that respond selectively at test time.

Gold-state generation further separates continuation ability from entry policy. $Q_{\rm Gold}$ forces every model to generate at the same 128 positive query states: GPT-4o leads this metric, while VITA-1.5 leads $Q_{\rm joint}^{\rm Ens}$ by responding to more opportunities. These metrics answer distinct questions and neither replaces fixed-time \emph{when} classification.

\subsection{Diagnostic Analysis}
\label{sec:diagnostic}

\noindent\textbf{Speaker attribution.}\quad
Speaker attribution degrades under visibility mismatch for most, but not all, systems. Against a 25\% uniform-random baseline, several models remain near or below chance after failures count as incorrect. Table~\ref{tab:visibility_results} shows gaps from $-4.58$ to $+10.06$ points across the naturally occurring strata. These gaps diagnose difficult grounding conditions without estimating a causal audio-visual conflict effect; supplementary results report macro-F1 and Wilson intervals for every system and visibility stratum.

\begin{table}[t]
\caption{Speaker-visible and naturally occurring visibility-mismatch accuracy. The gap is descriptive, not causal.}
\label{tab:visibility_results}
\centering
\setlength{\tabcolsep}{3pt}
{\small
\begin{tabular*}{\columnwidth}{@{\extracolsep{\fill}}lrrr}\toprule
Model & Visible & Mismatch & $\Delta_{\rm vis}$ \\\midrule
GPT-4o & 35.59 & 31.64 & +3.96 \\
Gemini 2.5 Pro & 40.81 & 34.18 & +6.63 \\
Gemini 2.5 Flash & 34.20 & 30.55 & +3.66 \\
Gemini 3 Flash & 49.39 & 40.00 & +9.39 \\
Gemini 3 Pro & 46.78 & 36.73 & +10.06 \\
Qwen3-Omni & 75.25 & 70.91 & +4.34 \\
Qwen3-Omni-Thinking & 68.35 & 63.27 & +5.08 \\
Qwen2.5-Omni & 41.28 & 38.91 & +2.37 \\
OmniVinci & 30.09 & 27.64 & +2.45 \\
VITA-1.5 & 31.42 & 36.00 & -4.58 \\
Baichuan-Omni-1.5 & 23.13 & 18.18 & +4.95 \\
\bottomrule\end{tabular*}}

\end{table}

\noindent\textbf{Turn-entry classification.}\quad
Turn-entry models exhibit distinct conservative and permissive policies. Because 128 of 200 items are positive, always predicting YES reaches 64.00\% accuracy but only 39.02 macro-F1. Reporting both metrics prevents prevalence from making a permissive policy appear strong; neither metric measures real-time latency.

\begin{table}[t]
\caption{When-only controls. Pause rules use $-35$ dBFS; Random uses seed 13; Whisper + pause combines punctuation with a 0.2-s pause.}
\label{tab:extension_when_baselines}
\centering
\setlength{\tabcolsep}{1pt}
\renewcommand{\arraystretch}{0.84}
{\small
\begin{tabular*}{\columnwidth}{@{\extracolsep{\fill}}lrrrrrr}\toprule
Baseline & Acc. & F1 & TP & FP & TN & FN \\\midrule
Always-NO & 36.00 & 26.47 & 0 & 0 & 72 & 128 \\
Always-YES & 64.00 & 39.02 & 128 & 72 & 0 & 0 \\
Majority & 64.00 & 39.02 & 128 & 72 & 0 & 0 \\
Random & 54.00 & 53.21 & 67 & 31 & 41 & 61 \\
Pause 0.2 s & 62.00 & 47.92 & 114 & 62 & 10 & 14 \\
Pause 0.5 s & 63.50 & 53.97 & 109 & 54 & 18 & 19 \\
Pause 1.0 s & 63.50 & 57.15 & 102 & 47 & 25 & 26 \\
Whisper punct. & 50.50 & 33.55 & 101 & 66 & 0 & 33 \\
Whisper + pause & 50.00 & 39.90 & 91 & 57 & 9 & 43 \\
\bottomrule\end{tabular*}}

\end{table}

Simple pause and transcript-endpoint controls do not solve fixed-time turn entry. Table~\ref{tab:extension_when_baselines} shows that the strongest fixed pause rule reaches 63.50\% accuracy and 57.15 macro-F1: it improves over always-YES macro-F1 but remains below the best OLM accuracy. We report every threshold to avoid selecting a pause duration on the test set.

The GPT-4o cascade illustrates how interface and decision policy jointly shape a score. It predicts \texttt{YES} on 49 of 200 items, including 39 true positives, and reaches 50.50\% accuracy, 49.84 macro-F1, and 30.47\% positive-opportunity coverage. Because the cascade converts truncated audio to text, these conservative-policy results characterize the adapter as well as the underlying model being evaluated.

Input-channel ablations expose interface dependence without identifying causal modality contributions. Table~\ref{tab:extension_modality_ablation} keeps the primary denominators fixed while comparing audio-only and video-only settings. Their model- and task-specific variation is interpreted as an interface effect.

The ablation shows model-specific channel sensitivity: audio-only inputs are stronger for most native omni models on \emph{who}, while video-only inputs generally reduce attribution and gold-state generation quality. GPT-4o shows a similarly large generation gap between its transcript-only and frames-only adapters. The complete per-model ablation table is provided in the supplementary material.

\noindent\textbf{Judge reliability.}\quad
Response judging is complete and auditable, but its members are not interchangeable. All three judges score every non-empty true-positive response, and the audit reports pairwise agreement, leave-one-judge-out rankings, and same-family removal. The primary leaderboard is defined only by this fixed, fully observed judge protocol.

\begin{table}[t]
\caption{Primary-judge robustness audit. Top: each judge against the complete ensemble and independent sanity controls; bottom: leave-one-out rank sensitivity and pairwise response agreement.}
\label{tab:extension_judge_audit}
\centering
\setlength{\tabcolsep}{1pt}
\renewcommand{\arraystretch}{0.78}
{\small
\begin{tabular*}{\columnwidth}{@{\extracolsep{\fill}}lrrrrrr}
\toprule
Audit & $\rho_r$ & MAE & $\tau_m$ & Shift & AUC & FPR \\\midrule
GPT-4o & 0.90 & 13.98 & 0.89 & 1 & 1.00 & 7.81 \\
Gemini & 0.76 & 17.25 & 0.75 & 5 & 0.93 & 10.16 \\
Qwen3 & 0.90 & 11.34 & 0.93 & 1 & 0.97 & 21.09 \\
\bottomrule\end{tabular*}
\vspace{2pt}
\begin{tabular*}{\columnwidth}{@{\extracolsep{\fill}}lrrlrrr}
\toprule
Removal & $\tau_m$ & Shift & Pair & $\rho_r$ & $\tau_r$ & MAE \\\midrule
w/o GPT & 0.93 & 1 & GPT / Gem. & 0.51 & 0.45 & 28.27 \\
w/o Gem. & 0.93 & 1 & GPT / Qwen & 0.73 & 0.64 & 17.22 \\
w/o Qwen & 0.96 & 1 & Gem. / Qwen & 0.59 & 0.51 & 24.06 \\
Same-family & 1.00 & 0 &  &  &  &  \\
\bottomrule

\end{tabular*}}

\end{table}

Individual judges correlate with the ensemble while retaining distinct lower-tier orderings. Table~\ref{tab:extension_judge_audit} reports response-level Spearman correlations of $0.76$--$0.90$ against the complete ensemble and model-level Kendall correlations of $0.75$--$0.93$. GPT-4o and Qwen3-Omni move any model by at most one rank, whereas Gemini~2.5 Pro moves one model by five ranks. On independent controls, each judge separates gold from shuffled continuations with AUC $0.93$--$1.00$; FPR denotes shuffled candidates scored at least 75.

Judge removal confirms that the ensemble ranking is not controlled by one evaluator. Each leave-one-out ranking has Kendall $\tau\geq0.93$ and changes any model by at most one position; same-family removal preserves the complete ordering. We retain the complete ensemble and report these shifts explicitly rather than claiming that family preference is absent. Full per-model differences appear in the supplementary material for independent inspection.

Score distributions still show evaluator differences. Pairwise response correlations range from $0.51$ to $0.73$, and pairwise MAE ranges from $17.22$ to $28.27$ points. Requiring all three scores fixes the response set, but averaging cannot remove rubric-interpretation differences. The ensemble is thus a reproducible surrogate rather than evidence of judge equivalence at the response level.

The coverage-adjusted generation results remain stable under resampling. GPT-4o retains the strongest conditional quality, whereas VITA-1.5 retains the strongest coverage and joint utility; Qwen3-Omni-Thinking combines the second-highest conditional quality with a joint score of $30.40$. Table~\ref{tab:appx_generation_coverage} reports the corresponding confidence intervals.

\noindent\textbf{Cross-axis model profiles.}\quad
The joint profile identifies distinct strengths among leading systems. Qwen3-Omni obtains $74.65\%$ speaker attribution but $42.59$ conditional response quality, so reliable speaker binding does not determine continuation quality. Qwen3-Omni-Thinking combines $67.65\%$ attribution with $67.77$ gold-state and $64.86$ conditional quality, but $46.88\%$ coverage lowers joint quality to $30.40$. VITA-1.5 has $32.05\%$ attribution, yet $65.50\%$ turn-entry accuracy and $88.28\%$ coverage produce the strongest joint score ($44.08$). These profiles expose each system's limiting component under the common evaluation protocol.

\noindent\textbf{Gold-state decomposition.}\quad
Forced generation separates continuation capacity from selective entry. GPT-4o remains strongest when every positive state is presented, with $Q_{\rm Gold}=77.15$, so its high conditional score is not merely an artifact of choosing easy responses. Qwen3-Omni-Thinking remains strong across all gold states ($67.77$) and selected responses ($64.86$). VITA-1.5 is similarly consistent between $Q_{\rm Gold}=48.37$ and $Q_{\rm Ens}=49.93$ while covering most opportunities. Together, the metrics distinguish strong generation, selective generation, and broad coverage across the same 128 dialogue states for every model.

\subsection{Robustness and Interpretation}
\label{sec:validity}

The results support a diagnostic, rather than scalar, interpretation of SocialOmni. Leadership reverses across all three axes: Qwen3-Omni leads speaker attribution, VITA-1.5 leads fixed-time turn entry and joint response quality, and GPT-4o leads gold-state and conditional generation. Moreover, $Q_{\rm Ens}$ measures produced-response quality, whereas $Q_{\rm joint}^{\rm Ens}$ incorporates missed opportunities. Their separation reveals whether progress comes from stronger continuation generation or a more effective entry policy.

Cross-channel comparisons reinforce the importance of speech-bearing evidence for continuation generation. Among configurations that return valid outputs, audio- or transcript-based inputs yield higher $Q_{\rm Gold}$ than video-only inputs: GPT-4o scores $78.97$ with transcript evidence versus $26.11$ with frames, Qwen3-Omni scores $59.51$ with audio versus $18.75$ with video, and VITA-1.5 scores $51.24$ versus $22.14$. These interface comparisons diagnose evidence sensitivity without asserting causal modality independence.

Metric triangulation exposes policy differences hidden by accuracy. The always-YES control obtains $64.00\%$ turn-entry accuracy because 128 of 200 items are positive, yet its macro-F1 is only $39.02$. OmniVinci reaches $62.03$ macro-F1 at similar accuracy, indicating stronger balance across entry and non-entry states. VITA-1.5 achieves the highest accuracy at $65.50\%$; together, the metrics separate balanced discrimination from opportunity capture in this split.

Bootstrap intervals support the response-policy distinction. VITA-1.5's positive-opportunity coverage interval is $[81.56,92.77]$, compared with $[23.16,38.92]$ for GPT-4o and $[38.45,55.48]$ for Qwen3-Omni-Thinking. Its joint-score advantage therefore reflects broad opportunity capture rather than a small set of favorable responses.

Visibility strata provide an independent grounding stress test. The visible-to-mismatch gap reaches $10.06$ points for Gemini~3 Pro and $9.39$ for Gemini~3 Flash, while Qwen3-Omni remains strongest in both strata at $75.25\%$ and $70.91\%$. The sign reversal for VITA-1.5 further shows that the partition is not a uniform difficulty rescaling. Reporting both strata therefore reveals model-specific grounding behavior hidden by aggregate speaker accuracy.

Independent judge checks support the central comparisons while exposing lower-tier sensitivity. Individual judges achieve model-level Kendall $\tau=0.75$--$0.93$ against the complete ensemble; the Gemini judge produces the largest single-judge shift of five ranks, but every leave-one-out ensemble changes models by at most one position. Together with fixed-denominator baselines, visibility strata, and uncertainty intervals, these audits triangulate the benchmark's principal comparisons across evidence and scoring views.

The control suite verifies that stable ranking is accompanied by basic response discrimination. Across 236 positive and 256 negative control candidates, the fixed three-judge reference-aware mean reaches gold-versus-shuffled AUC $0.99$ with a $74.67$-point paired margin. This result supports the disclosed ensemble for comparative diagnosis, while the complete per-judge distributions remain available for independent reanalysis of judge behavior.

\section{Conclusion}
\label{sec:conclusion}

SocialOmni reframes omni-model evaluation around the conversational turn: grounding \emph{who} is speaking, deciding \emph{when} a designated participant should enter, and determining \emph{how} that participant should respond. Its 2,200-item benchmark, fixed-prefix protocol, gold-state generation track, and coverage-aware response metrics turn these coupled decisions into an auditable diagnostic profile. The accompanying visibility, modality, baseline, and judge-sensitivity analyses make the profile interpretable rather than reducing social interaction to a single opaque score.

Evaluation of 11 OLMs yields three actionable findings. First, leadership changes across speaker attribution, turn entry, and response generation, so static understanding accuracy does not determine interaction quality. Second, gold-state and coverage-adjusted scores separate continuation ability from the policy that decides whether to speak. Third, visibility strata and input-channel ablations expose dependence on scene evidence and interfaces, while the judge audit quantifies open-ended evaluation sensitivity. Together, these results make SocialOmni a practical foundation for socially interactive omni-model development and comparison.

Because each score maps to a distinct failure mode, the resulting profile gives developers concrete targets for improving speaker grounding, entry policy, and response quality while keeping progress across the three interaction stages separately measurable under a common evaluation contract.

\FloatBarrier
\section*{Reproducibility Statement}
The benchmark construction and fixed-prefix evaluation protocol are described in Section~\ref{sec:method}. Appendix~\ref{sec:method_appendix} provides annotation procedures, model and inference details, response-scoring rules, and prompt templates. Appendix~\ref{sec:results_appendix} reports uncertainty, failure accounting, and judge-sensitivity analyses to support interpretation and reproduction of the reported results.

\section*{AI Use Statement}
In preparing this manuscript, we used an AI assistant for language polishing, manuscript formatting, and searching for related work. AI assistance was not used to generate research ideas, formulate hypotheses, design the methodology or experiments, or interpret the results. The authors take responsibility for the final content of the paper, including its claims and conclusions. Separately, the generative models evaluated as benchmark participants and used as response judges are components of the experimental protocol, as described in the paper and Appendix~\ref{appx:multi_ref}.

\clearpage
\bibliographystyle{plainnat}
\bibliography{socialomni}

\clearpage
\appendix
\setcounter{secnumdepth}{2}
\renewcommand{\thefigure}{S\arabic{figure}}
\renewcommand{\thetable}{S\arabic{table}}
\renewcommand{\thealgorithm}{S\arabic{algorithm}}
\setcounter{figure}{0}
\setcounter{table}{0}
\setcounter{algorithm}{0}
\raggedbottom
\newcommand{\suppentry}[2]{%
  \begin{tabularx}{\linewidth}{@{}>{\raggedright\arraybackslash}Xr@{}}
  \supplink{#1}{#2}\dotfill & \supplink{\mbox{\pageref{#2}}}{#2}
  \end{tabularx}\par}

\begin{minipage}{\textwidth}
\centering
{\LARGE\bfseries Supplementary Material}\par
\vspace{2pt}
{\large\bfseries Contents}\par
\vspace{5pt}
\hrule
\vspace{6pt}
\begin{minipage}[t]{\textwidth}
\normalsize
\suppentry{\textbf{A. Additional Method Details}}{sec:method_appendix}
\smallskip
\suppentry{A.1 Human Annotation and Agreement}{appx:iaa}
\suppentry{A.2 Domain Definitions and Size Rationale}{appx:category_details}
\suppentry{A.3 Speaker-Visibility Stratification}{appx:boundary_cases}
\suppentry{A.4 Perception Option Balance}{appx:option_balance}
\suppentry{A.5 Fixed-Time Turn-Entry Protocol}{appx:when_protocol}
\suppentry{A.6 Response Scoring and Coverage}{appx:response_scoring}
\suppentry{A.7 Judge Sensitivity}{appx:multi_ref}
\suppentry{A.8 Model and Inference Details}{appx:model_details}
\suppentry{A.9 Reproducibility and Release Scope}{appx:reproducibility}
\suppentry{A.10 Prompt Templates and Parsing Rules}{appx:prompts}
\end{minipage}\hfill
\medskip
\begin{minipage}[t]{\textwidth}
\normalsize
\suppentry{\textbf{B. Additional Results}}{sec:results_appendix}
\smallskip
\suppentry{B.1 Classification Reliability}{appx:reliability_tables}
\suppentry{B.2 Response-Evaluation Details}{appx:coverage_results}
\suppentry{B.3 Blind Three-Annotator Calibration}{appx:human_feedback}
\suppentry{B.4 Representative Failure Cases}{appx:failure_cases}
\end{minipage}
\medskip
\begin{minipage}[t]{\textwidth}
\normalsize
\suppentry{\textbf{C. Reliability, Bias, and Scope}}{appx:reliability_scope}
\smallskip
\suppentry{C.1 Three-Annotator Verification}{appx:human_annotation_revision}
\suppentry{C.2 Target Identity and Visibility}{appx:identity_consistency_revision}
\suppentry{C.3 Judge Alignment and Bias}{appx:judge_revision}
\suppentry{C.4 Benchmark Scope}{appx:scope_limitations}
\end{minipage}
\vspace{5pt}
\hrule
\end{minipage}

\clearpage
\suppanchor{sec:method_appendix}
\section{Additional Method Details}
\label{sec:method_appendix}

\suppanchor{appx:iaa}
\subsection{Human Annotation and Agreement}
\label{appx:iaa}

Every benchmark item is independently checked by three trained human annotators before adjudication. For \emph{who}, annotators verify the active speaker, spoken content, four answer options, and speaker-visibility stratum. For \emph{when}, they verify the target-participant description, query time, and YES/NO label. For \emph{how}, they verify the target role and write the reference continuation. Disagreements are cross-reviewed and resolved by a senior annotator. The reported pre-adjudication raw agreement is 94.2\% for perception annotations and 91.8\% for interaction-generation annotations. We distinguish these construction-stage figures from the adjudicated gold labels distributed with the benchmark; chance-corrected agreement cannot be reconstructed from the public adjudicated labels alone.

\suppanchor{appx:category_details}
\subsection{Domain Definitions and Size Rationale}
\label{appx:category_details}

The 15 dialogue categories are grouped into four broad domains: entertainment, professional, daily life, and narrative. Entertainment covers interactive media such as talk shows and podcasts; professional covers interviews, debates, and task-oriented settings; daily life covers everyday conversations; narrative covers scripted movie and drama scenes. These labels support stratification, but the 200-item interaction-generation split is not large enough for stable per-category leaderboards.

\label{appx:gen_size_rationale}
The interaction-generation split is a manually curated diagnostic set. Scaling open-ended continuation indiscriminately can add ambiguity in target-role identification, query-state selection, and reference quality. Each retained item therefore requires a clearly described target participant, an agreed YES/NO entry label at the query time, sufficient audio-visual context, and a human-written continuation. This quality criterion motivates the 200-item core design; it does not imply that 200 items exhaust the diversity of social interaction.

\suppanchor{appx:boundary_cases}
\subsection{Speaker-Visibility Stratification}
\label{appx:boundary_cases}

The perception split contains 1,725 speaker-visible and 275 speaker-visibility-mismatch items. Labels are based on whether the active speaker can be attributed from the visible scene at the query time. Mismatch cases include off-screen speech, reaction shots, partial occlusion, and weak identity cues. They are naturally sampled rather than created by replacing audio or pairing unrelated clips. The visibility gap is therefore descriptive and may also reflect shot type, composition, or domain.

\suppanchor{appx:option_balance}
\subsection{Perception Option Balance}
\label{appx:option_balance}

The correct-option distribution over 2,000 \emph{who} items is A:~569, B:~561, C:~453, and D:~417. We retain this natural imbalance and report macro-F1 in addition to accuracy. A non-parsable output counts as incorrect and as a false negative for its gold class; because it has no predicted class, it contributes no class-specific false positive.

\suppanchor{appx:when_protocol}
\subsection{Fixed-Time Turn-Entry Protocol}
\label{appx:when_protocol}

Each interaction item provides an annotated query time $t_i$, target participant $X_i$, and binary label $y_i$. The target is identified by a concrete visual description such as clothing, location, or a visible action. Every model receives evidence bounded by the same prefix ending at $t_i$ and answers whether $X_i$ should start a substantive turn; disclosed adapters preserve this temporal boundary when an API lacks a joint media endpoint. The task does not repeatedly query earlier prefixes, estimate a first-YES timestamp, or include inference latency in the score.

This fixed-time formulation avoids imposing a universal asymmetric timing window on heterogeneous conversations. It also makes the protocol interface-agnostic across APIs and open-source checkpoints. Native streaming evaluation, persistent dialogue state, first-token latency, and acceptable-entry intervals remain separate extensions.

\suppanchor{appx:response_scoring}
\subsection{Response Scoring and Coverage}
\label{appx:response_scoring}

For the policy-conditioned evaluation, a response is generated on a gold-positive item only when the model predicts \texttt{YES}. Only non-empty true-positive responses enter conditional quality $Q_{\rm Ens}$. Missed opportunities and empty responses are excluded from this conditional mean but contribute zero to $Q_{\rm joint}^{\rm Ens}$ over the fixed 128 positive opportunities. Conditional quality is always reported with positive-opportunity coverage and the joint score. In the separate gold-state evaluation, $Q_{\rm Gold}$ forces generation at all 128 positive query states regardless of the model's entry prediction; failed or missing responses contribute zero to that fixed-denominator mean.

Each eligible non-empty response is scored independently by GPT-4o, Gemini~2.5 Pro, and Qwen3-Omni on $\{0,25,50,75,100\}$. The scale is interpreted as follows:
\begin{itemize}[leftmargin=*,nosep]
    \item 100: contextually grounded, role-consistent, coherent, and pragmatically natural;
    \item 75: appropriate overall, with minor genericness or missing detail;
    \item 50: partly relevant but with a noticeable grounding, role, or coherence problem;
    \item 25: weakly relevant, contradictory, or pragmatically inappropriate;
    \item 0: unrelated, wrong-role, or incompatible with the dialogue state.
\end{itemize}
Judges receive the manually verified dialogue context ending before the annotated continuation, the continuation itself as a reference, and the evaluated response. The reference is withheld from evaluated models. The rubric explicitly permits valid paraphrases and alternative continuations rather than treating lexical similarity as the objective.

Every non-empty true-positive response requires all three judge scores, whose arithmetic mean defines its ensemble quality. Failed judge calls are retried rather than averaged over a smaller pool. The evaluation artifact stores raw outputs, parsed scores, retries, model versions, and invocation dates. Pairwise agreement, leave-one-judge-out rankings, and same-family removal test whether the main conclusions depend on one judge.

\suppanchor{appx:multi_ref}
\subsection{Judge Sensitivity}
\label{appx:multi_ref}

Single-judge comparisons and leave-one-judge-out checks use different rankings: the former compare one judge with the complete ensemble, whereas the latter compare the remaining two-judge mean with that ensemble. Thus the five-position shift for the Gemini judge is compatible with a maximum one-position shift in each leave-one-out ranking. Same-family removal preserves the complete ordering. These checks reveal sensitivity within the fixed reference-aware protocol but cannot eliminate reference-style or model-family bias.

\suppanchor{appx:model_details}
\subsection{Model and Inference Details}
\label{appx:model_details}

Commercial model identifiers cover GPT-4o~\citep{hurst2024gpt}, Gemini~2.5 Pro/Flash~\citep{comanici2025gemini}, and Gemini~3 Flash/Pro Preview~\citep{google2025gemini3}. Open-source systems use Qwen3-Omni Instruct/Thinking~\citep{Qwen3-Omni}, Qwen2.5-Omni~\citep{xu2025qwen25omni}, OmniVinci~\citep{ye2025omnivinci}, VITA-1.5~\citep{fu2025vita}, and Baichuan-Omni-1.5~\citep{li2025baichuan}. For GPT-4o, the already truncated query prefix is converted to 16-kHz mono audio and transcribed by \texttt{gpt-4o-transcribe}; GPT-4o receives this machine transcript together with sampled prefix frames. The Gemini relay used in this evaluation accepts sampled frames through its OpenAI-compatible endpoint but does not expose the media audio track. We therefore retain Gemini as an explicitly interface-limited baseline rather than labeling it native audio-video inference. No adapter receives content after the query time, and evaluated models never receive the human reference continuation. Decoding uses temperature 0.3 and top-$p$ 1.0 where exposed; judge decoding is deterministic where supported. Exact identifiers, invocation dates, request envelopes, preprocessing, timeout, retry settings, and interface metadata are retained with the versioned evaluation snapshot.

\suppanchor{appx:reproducibility}
\subsection{Reproducibility and Release Scope}
\label{appx:reproducibility}

The supplementary package provides adjudicated annotations, frozen evaluation results, and offline analysis scripts. The original video media are not included in the package. The released evaluation manifest fixes 2,000 perception items and 200 interaction-generation items (\texttt{video\_0001}--\texttt{video\_0200}), matching every denominator reported in this paper. The public annotations include item IDs, media paths, labels, dialogue metadata, and the visible/mismatch field used in our stratified analysis. The public code repository is available at \url{https://github.com/MAC-AutoML/SocialOmni}, and the public dataset, including video media, is available at \url{https://huggingface.co/datasets/alexisty/SocialOmni}. These public resources are distinct from the frozen evaluation snapshot described below.

We distinguish the public dataset artifact from the versioned evaluation snapshot. The latter retains raw model text, parsed predictions, parse failures, response presence, judge scores, interface type, and run metadata; tables and figures are generated from that single snapshot. Automated checks verify fixed item IDs, model means, score support, and coverage identities. Source hashes, original download dates, and per-source licensing fields are not currently present in the public dataset repository, so we do not claim that those provenance fields can be reconstructed from the release.

\suppanchor{appx:prompts}
\subsection{Prompt Templates and Parsing Rules}
\label{appx:prompts}

\begin{AIbox}[breakable=false]{\textbf{SocialOmni Prompt Cards}}
\textbf{Who.}\par
\texttt{Identify who says what in the marked interval. Answer only A, B, C, or D.}\par\medskip
\textbf{When.}\par
\texttt{You are an external dialogue observer. The target participant is [visual description]. Based only on the video and audio observed up to the current timestamp, should this participant begin a substantive speaking turn now? A means YES and B means NO. Answer only A or B.}\par\medskip
\textbf{How.}\par
\texttt{Write the next utterance that the target participant should say based only on the observed audio-visual prefix. Do not write speech for another participant. Output only the utterance.}
\end{AIbox}

For \emph{who}, only one letter in \{A,B,C,D\} is accepted. For \emph{when}, option A denotes YES and option B denotes NO; normalized unambiguous A/B or YES/NO outputs are accepted and mapped to these labels. Other outputs count as incorrect. For policy-conditioned \emph{how}, an empty output is excluded from conditional quality and contributes zero to the joint score; in the gold-state evaluation, it contributes zero to the fixed-denominator gold-state score.

\begin{figure*}[t]
  \centering
  \includegraphics[width=0.96\textwidth]{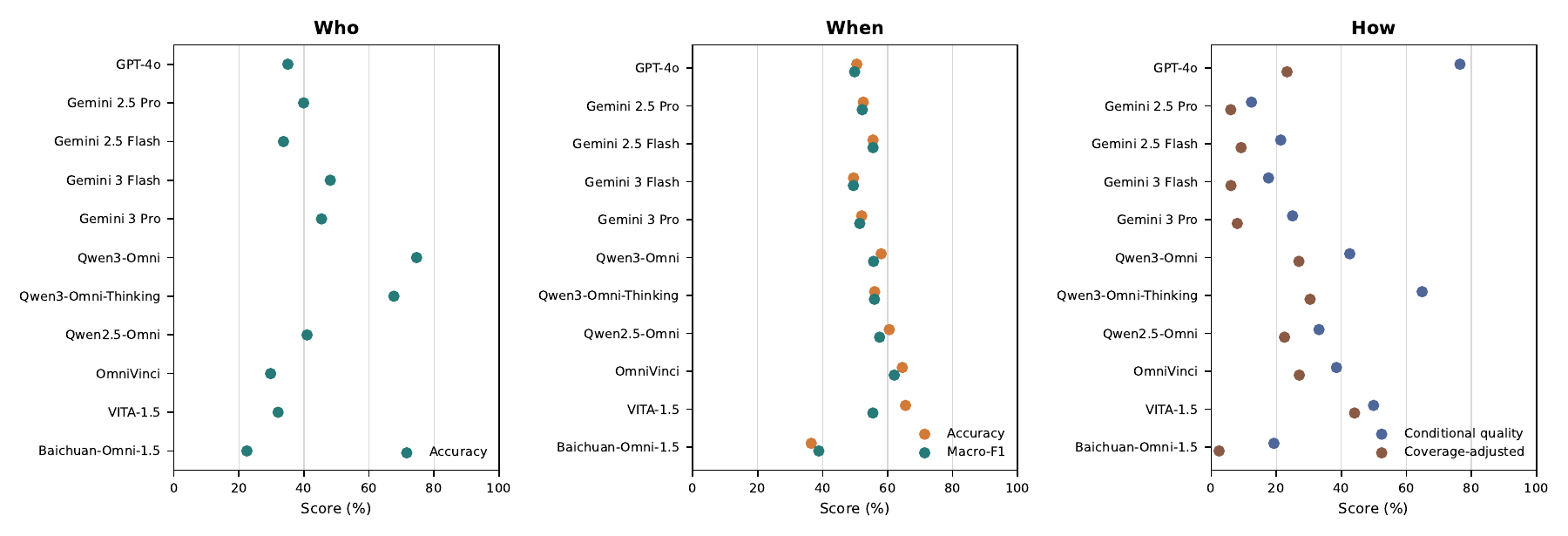}
  \caption{Model-level diagnostic profiles. \emph{Who} reports strict-denominator accuracy; \emph{when} contrasts accuracy and macro-F1; \emph{how} contrasts conditional ensemble quality with coverage-adjusted quality.}
  \label{fig:appx_diagnostic_profiles}
\end{figure*}

\suppanchor{sec:results_appendix}
\section{Additional Results}
\label{sec:results_appendix}

Figure~\ref{fig:appx_diagnostic_profiles} summarizes the principal model-level profiles behind the main tables. The point plots preserve the native metric scales and expose rank changes across speaker attribution, turn-entry classification, conditional response quality, and coverage adjustment.

\suppanchor{appx:reliability_tables}
\subsection{Classification Reliability}
\label{appx:reliability_tables}

Table~\ref{tab:perception_f1_ci} reports macro-F1 and Wilson intervals for both classification tasks. Request and parse failures are listed separately in Table~\ref{tab:failure_accounting}, remain in the strict denominators, and enter the gold-class error terms used for recall and macro-F1. Positive-class precision, recall, and F1 are released in the canonical summary.

\begin{table}[t]
\caption{Strict classification results. CIs are Wilson 95\% intervals for accuracy; When TP/FP/TN/FN counts cover parsed outputs, while failures remain in the 200-item metric denominator and are itemized separately.}
\label{tab:perception_f1_ci}
\centering
{\small
\begin{tabular*}{\linewidth}{@{\extracolsep{\fill}}lrrl@{}}
\toprule
Model & Who Acc. & Who F1 & Who CI \\
\midrule
GPT-4o & 35.05 & 31.18 & [32.99, 37.17] \\
Gemini 2.5 Pro & 39.90 & 39.62 & [37.78, 42.06] \\
Gemini 2.5 Flash & 33.70 & 33.64 & [31.66, 35.80] \\
Gemini 3 Flash & 48.10 & 47.88 & [45.92, 50.29] \\
Gemini 3 Pro & 45.40 & 45.20 & [43.23, 47.59] \\
Qwen3-Omni & 74.65 & 74.34 & [72.70, 76.51] \\
Qwen3-Omni-Thinking & 67.65 & 67.94 & [65.57, 69.66] \\
Qwen2.5-Omni & 40.95 & 39.50 & [38.81, 43.12] \\
OmniVinci & 29.75 & 30.00 & [27.79, 31.79] \\
VITA-1.5 & 32.05 & 31.38 & [30.04, 34.13] \\
Baichuan-Omni-1.5 & 22.45 & 11.60 & [20.68, 24.33] \\
\bottomrule
\end{tabular*}
\par\medskip
\begin{tabular*}{\linewidth}{@{\extracolsep{\fill}}lrrlr@{}}
\toprule
Model & When Acc. & When F1 & When CI & TP/FP/TN/FN \\
\midrule
GPT-4o & 50.50 & 49.84 & [43.63, 57.35] & 39/10/62/89 \\
Gemini 2.5 Pro & 52.50 & 52.20 & [45.60, 59.31] & 62/28/43/66 \\
Gemini 2.5 Flash & 55.50 & 55.50 & [48.57, 62.22] & 55/16/56/73 \\
Gemini 3 Flash & 49.50 & 49.47 & [42.65, 56.37] & 44/16/55/84 \\
Gemini 3 Pro & 52.00 & 51.41 & [45.10, 58.82] & 41/9/63/87 \\
Qwen3-Omni & 58.00 & 55.65 & [51.07, 64.63] & 81/37/35/47 \\
Qwen3-Omni-Thinking & 56.00 & 55.93 & [49.07, 62.70] & 60/20/52/68 \\
Qwen2.5-Omni & 60.50 & 57.52 & [53.59, 67.02] & 87/38/34/41 \\
OmniVinci & 64.50 & 62.03 & [57.65, 70.80] & 90/33/39/38 \\
VITA-1.5 & 65.50 & 55.45 & [58.68, 71.74] & 113/54/18/15 \\
Baichuan-Omni-1.5 & 36.50 & 38.82 & [30.14, 43.37] & 16/5/57/74 \\
\bottomrule
\end{tabular*}}

\end{table}

\suppanchor{appx:coverage_results}
\subsection{Response-Evaluation Details}
\label{appx:coverage_results}

The detailed response-level release retains every component score and satisfies $Q_{\rm joint}^{\rm Ens}=Q_{\rm Ens}\times\mathrm{Cov}^{+}/100$ when coverage is represented as a percentage. The remaining tables provide classification reliability, strict failure accounting, per-model family sensitivity, and additional sanity-control details.

\begin{table*}[t]
\caption{Three-judge conditional response quality and positive-opportunity adjustment.}
\label{tab:appx_generation_coverage}
\centering
{\small
\begin{tabular*}{\textwidth}{@{\extracolsep{\fill}}lrrrrr}
\toprule
Model & $Q_{\rm Ens}$ & Quality CI & Cov.$^{+}$ & Coverage CI & $Q_{\rm joint}^{\rm Ens}$ \\\midrule
GPT-4o & 76.50 & [67.31, 85.26] & 30.47 & [23.16, 38.92] & 23.31 \\
Gemini 2.5 Pro & 12.37 & [8.20, 17.07] & 48.44 & [39.95, 57.01] & 5.99 \\
Gemini 2.5 Flash & 21.36 & [15.30, 27.88] & 42.97 & [34.72, 51.63] & 9.18 \\
Gemini 3 Flash & 17.61 & [10.98, 25.00] & 34.38 & [26.71, 42.95] & 6.05 \\
Gemini 3 Pro & 25.00 & [18.29, 32.11] & 32.03 & [24.57, 40.54] & 8.01 \\
Qwen3-Omni & 42.59 & [35.39, 49.59] & 63.28 & [54.66, 71.13] & 26.95 \\
Qwen3-Omni-Thinking & 64.86 & [57.50, 72.36] & 46.88 & [38.45, 55.48] & 30.40 \\
Qwen2.5-Omni & 33.14 & [29.60, 36.88] & 67.97 & [59.46, 75.43] & 22.53 \\
OmniVinci & 38.52 & [31.94, 45.28] & 70.31 & [61.90, 77.54] & 27.08 \\
VITA-1.5 & 49.93 & [43.88, 55.75] & 88.28 & [81.56, 92.77] & 44.08 \\
Baichuan-Omni-1.5 & 19.27 & [10.42, 29.69] & 12.50 & [7.84, 19.34] & 2.41 \\
\bottomrule\end{tabular*}}

\end{table*}

\begin{table}[t]
\caption{Bootstrap 95\% intervals for conditional response quality and Wilson 95\% intervals for positive-opportunity coverage. Point estimates are in Table~\ref{tab:main_triplet}.}
\label{tab:generation_uncertainty}
\centering
\setlength{\tabcolsep}{2.5pt}
{\small
\begin{tabular*}{\columnwidth}{@{\extracolsep{\fill}}lrr}
\toprule Model & $Q_{\rm Ens}$ CI & Cov.$^{+}$ CI \\\midrule
GPT-4o & [67.3,85.3] & [23.2,38.9] \\
Gemini 2.5 Pro & [8.2,17.1] & [40.0,57.0] \\
Gemini 2.5 Flash & [15.3,27.9] & [34.7,51.6] \\
Gemini 3 Flash & [11.0,25.0] & [26.7,43.0] \\
Gemini 3 Pro & [18.3,32.1] & [24.6,40.5] \\
Qwen3-Omni & [35.4,49.6] & [54.7,71.1] \\
Qwen3-Omni-Thinking & [57.5,72.4] & [38.4,55.5] \\
Qwen2.5-Omni & [29.6,36.9] & [59.5,75.4] \\
OmniVinci & [31.9,45.3] & [61.9,77.5] \\
VITA-1.5 & [43.9,55.8] & [81.6,92.8] \\
Baichuan-Omni-1.5 & [10.4,29.7] & [7.8,19.3] \\
\bottomrule\end{tabular*}}

\end{table}

\IfFileExists{tables/extension_how_gold_main.tex}{\begin{table*}[t]
\caption{Complete gold-state generation results. $Q_{\rm Gold}$ is the three-judge reference-aware mean over the fixed 128 positive states; failed or missing responses score zero. CI is a 95\% item bootstrap interval.}
\label{tab:extension_how_gold}
\centering
\setlength{\tabcolsep}{0.6pt}
{\small
\begin{minipage}[t]{\linewidth}\centering
\begin{tabular}{@{}lcrrl@{}}\toprule Model & Interface & Valid & $Q_{\rm Gold}$ & 95\% CI \\\midrule
Baichuan-Omni-1.5 & audio-video & 128 (0 fail) & 18.36 & [14.32, 22.85] \\
Gemini 2.5 Flash & visual-only & 128 (0 fail) & 21.35 & [17.64, 25.26] \\
Gemini 2.5 Pro & visual-only & 128 (0 fail) & 15.62 & [11.91, 19.66] \\
Gemini 3 Flash & visual-only & 128 (0 fail) & 13.93 & [10.68, 17.32] \\
Gemini 3 Pro & visual-only & 125 (3 fail) & 21.55 & [17.97, 25.13] \\
GPT-4o & cascade & 127 (1 fail) & 77.15 & [72.72, 81.38] \\
OmniVinci & audio-video & 128 (0 fail) & 40.82 & [35.87, 46.22] \\
Qwen2.5-Omni & audio-video & 128 (0 fail) & 36.00 & [32.75, 39.32] \\
Qwen3-Omni & audio-video & 128 (0 fail) & 44.47 & [38.87, 50.07] \\
Qwen3-Omni-Thinking & audio-video & 128 (0 fail) & 67.77 & [62.69, 72.85] \\
VITA-1.5 & audio-video & 128 (0 fail) & 48.37 & [42.58, 54.04] \\
\bottomrule\end{tabular}\end{minipage}\par\medskip
\begin{minipage}[t]{\linewidth}\centering
\begin{tabular}{@{}lcrl@{}}\toprule Model & Variant & $Q_{\rm Gold}$ & 95\% CI \\\midrule
Baichuan-Omni-1.5 & audio-only & 8.98 & [6.45, 11.79] \\
Baichuan-Omni-1.5 & video-only & 16.73 & [12.89, 20.57] \\
GPT-4o & frames-only & 26.11 & [21.81, 30.47] \\
GPT-4o & transcript-only & 78.97 & [74.41, 83.33] \\
OmniVinci & audio-only & 42.12 & [36.91, 47.46] \\
OmniVinci & video-only & 20.83 & [17.25, 24.67] \\
Qwen2.5-Omni & audio-only & 41.08 & [35.94, 45.96] \\
Qwen2.5-Omni & video-only & 1.95 & [1.11, 2.93] \\
Qwen3-Omni & audio-only & 59.51 & [53.65, 65.30] \\
Qwen3-Omni & video-only & 18.75 & [15.36, 22.40] \\
Qwen3-Omni-Thinking & audio-only & 66.73 & [61.46, 71.94] \\
Qwen3-Omni-Thinking & video-only & 26.63 & [22.20, 31.12] \\
VITA-1.5 & audio-only & 51.24 & [45.77, 56.90] \\
VITA-1.5 & video-only & 22.14 & [17.71, 26.82] \\
\bottomrule\end{tabular}\end{minipage}}

\end{table*}
}{}

Table~\ref{tab:extension_how_gold} reports the complete fixed-state generation audit. Every model is prompted at the same 128 gold-positive query states, and missing generations remain in the denominator with zero quality. The modality rows are diagnostic interface variants rather than additional leaderboard entries; the audio-video or disclosed adapter row supplies $Q_{\rm Gold}$ in the primary table.
\IfFileExists{tables/extension_modality_ablation.tex}{\begin{table*}[t]
\caption{Input-channel ablations. A/V denotes audio-only/video-only; T/F denotes GPT-4o transcript-only/frames-only adapters. Primary-interface scores remain in Table~\ref{tab:main_triplet}.}
\label{tab:extension_modality_ablation}
\centering
\setlength{\tabcolsep}{2.5pt}
{\small
\begin{tabular*}{\textwidth}{@{\extracolsep{\fill}}llrrrrrr}\toprule
Model & Pair & \multicolumn{2}{c}{Who} & \multicolumn{2}{c}{When} & \multicolumn{2}{c}{$Q_{\rm Gold}$} \\
 & & A/T & V/F & A/T & V/F & A/T & V/F \\\midrule
GPT-4o & T/F & 35.75 & 30.50 & 56.50 & 36.00 & 78.97 & 26.11 \\
Qwen3-Omni & A/V & 71.45 & 37.05 & 33.00 & 30.00 & 59.51 & 18.75 \\
Qwen3-Omni-Thinking & A/V & 47.80 & 34.80 & 0.00 & 0.00 & 66.73 & 26.63 \\
Qwen2.5-Omni & A/V & 41.45 & 33.05 & 24.50 & 20.50 & 41.08 & 1.95 \\
OmniVinci & A/V & 34.35 & 28.30 & 21.50 & 20.00 & 42.12 & 20.83 \\
VITA-1.5 & A/V & 35.15 & 31.95 & 60.50 & 66.00 & 51.24 & 22.14 \\
Baichuan-Omni-1.5 & A/V & 23.40 & 27.15 & 39.50 & 7.00 & 8.98 & 16.73 \\
\bottomrule\end{tabular*}}

\end{table*}
}{}
\begin{table}[t]
\caption{Judge-family sensitivity over eligible responses. N.A. denotes no eligible response; shift is the maximum leave-one-judge-out rank change.}
\label{tab:judge_sensitivity}
\centering
\setlength{\tabcolsep}{1pt}
{\small
\begin{tabular}{@{}lrrrrr@{}}\toprule
Model & Resp. & Full & No-fam. & $\Delta$ & Shift \\\midrule
GPT-4o & 39 & 76.50 & 71.79 & -4.70 & 0 \\
Gemini 2.5 Pro & 62 & 12.37 & 13.91 & +1.55 & 0 \\
Gemini 2.5 Flash & 55 & 21.36 & 22.27 & +0.91 & 1 \\
Gemini 3 Flash & 44 & 17.61 & 16.76 & -0.85 & 1 \\
Gemini 3 Pro & 41 & 25.00 & 26.22 & +1.22 & 1 \\
Qwen3-Omni & 81 & 42.59 & 42.13 & -0.46 & 0 \\
Qwen3-Omni-Thinking & 60 & 64.86 & 63.54 & -1.32 & 0 \\
Qwen2.5-Omni & 87 & 33.14 & 28.02 & -5.12 & 1 \\
OmniVinci & 90 & 38.52 & 38.52 & +0.00 & 1 \\
VITA-1.5 & 113 & 49.93 & 49.93 & +0.00 & 0 \\
Baichuan-Omni-1.5 & 16 & 19.27 & 19.27 & +0.00 & 1 \\
\bottomrule\end{tabular}}

\end{table}

\begin{table}[t]
\caption{Strict-denominator failure accounting. Request/runtime and parse failures are disjoint; all remain in the fixed denominators.}
\label{tab:failure_accounting}
\centering
\setlength{\tabcolsep}{1pt}
{\small
\begin{tabular*}{\columnwidth}{@{\extracolsep{\fill}}lrrrrr@{}}
\toprule Model & W req. & W parse & T req. & T parse & Scored \\\midrule
GPT-4o & 12 & 1 & 0 & 0 & 39 \\
Gemini 2.5 Pro & 6 & 0 & 1 & 0 & 62 \\
Gemini 2.5 Flash & 3 & 2 & 0 & 0 & 55 \\
Gemini 3 Flash & 6 & 1 & 1 & 0 & 44 \\
Gemini 3 Pro & 10 & 0 & 0 & 0 & 41 \\
Qwen3-Omni & 0 & 0 & 0 & 0 & 81 \\
Qwen3-Omni-Thinking & 0 & 27 & 0 & 0 & 60 \\
Qwen2.5-Omni & 0 & 0 & 0 & 0 & 87 \\
OmniVinci & 0 & 263 & 0 & 0 & 90 \\
VITA-1.5 & 0 & 1 & 0 & 0 & 113 \\
Baichuan-Omni-1.5 & 0 & 0 & 0 & 48 & 16 \\
\bottomrule\end{tabular*}}

\end{table}

\IfFileExists{tables/extension_supplementary_details.tex}{\begin{table}[t]
\caption{Supplementary final-extension inventory and selected sanity-control results. Detailed CSV/JSON snapshots remain the source of record.}
\label{tab:extension_supplementary_details}
\centering
\setlength{\tabcolsep}{0.5pt}
{\small
\begin{tabular}{lr}\toprule
Snapshot family & Rows \\\midrule
How@Gold groups & 25 \\
Modality groups & 32 \\
When baselines & 9 \\
Sanity judge groups & 16 \\
\bottomrule\end{tabular}}
\vspace{2pt}
{\small\begin{tabular}{lrrrr}\toprule
Sanity setting & Pos. & Neg. & G/S AUC & G--S \\\midrule
Primary / ref.-aware & 236 & 256 & 0.99 & 74.67 \\
Primary / ref.-free & 236 & 256 & 0.53 & 1.69 \\
Held-out / ref.-aware & 236 & 256 & 1.00 & 68.46 \\
Held-out / ref.-free & 236 & 256 & 0.46 & -1.37 \\
GPT-4o / ref.-aware & 236 & 256 & 1.00 & 75.00 \\
GPT-4o / ref.-free & 236 & 256 & 0.51 & 0.98 \\
Gemini / ref.-aware & 236 & 256 & 0.93 & 79.88 \\
Gemini / ref.-free & 236 & 256 & 0.52 & 1.37 \\
Qwen3 / ref.-aware & 236 & 256 & 0.97 & 69.14 \\
Qwen3 / ref.-free & 236 & 256 & 0.52 & 2.73 \\
\bottomrule\end{tabular}}

\end{table}
}{}

Strict failure accounting treats interface compliance as part of measured performance. Request or runtime failures are uncommon, while invalid outputs concentrate in a few open systems: OmniVinci has 263 unparsable \emph{who} outputs, Qwen3-Omni-Thinking has 27, and Baichuan-Omni-1.5 has 48 unparsable \emph{when} outputs. All such cases remain in the fixed denominators; the table reports them separately rather than silently removing them.

The control audit exposes a sharp reference-sensitivity boundary. Reference-aware controls separate verified from shuffled continuations almost perfectly, whereas archived reference-free controls do not preserve that ordering. We therefore retain the fixed reference-aware ensemble for the primary score and avoid interpreting judge agreement as human equivalence.

\FloatBarrier
\suppanchor{appx:human_feedback}
\subsection{Blind Three-Annotator Calibration}
\label{appx:human_feedback}

The blind three-annotator calibration assesses response quality without redefining the primary leaderboard. Three annotators, blinded to source-model identity, independently score 200 final responses from ten systems on the same $\{0,25,50,75,100\}$ support as the ensemble; the analysis uses their per-response mean. GPT-4o is absent from this frozen subset, so the calibration does not define a complete model ranking.

Human and ensemble assessments agree most clearly at the capability-tier level. Response-level Pearson correlation is $0.619$ (95\% bootstrap CI $[0.492,0.726]$), model-level Pearson correlation is $0.817$, and eight of ten systems share the same descriptive Poor/Moderate/Strong tier (80\%; $\kappa=0.655$). The top two systems preserve their order, while most rank changes occur in the compressed low-quality region. Human ratings are internally reliable (ICC(2,$k$)$=0.925$; 94\% within 25 points). VITA-1.5 is the principal model-specific discrepancy, supporting continued human auditing rather than a claim of judge equivalence. Table~\ref{tab:human_feedback} gives all model-level means.

\begin{table}[!t]
\caption{Blind three-annotator calibration on 200 final responses from ten systems. Annotator and LLM values are three-rater and three-judge means; tiers are Poor ($<25$), Moderate ($25$--$49.99$), and Strong ($\geq50$). GPT-4o is absent from this fixed subset.}
\label{tab:human_feedback}
\centering
\setlength{\tabcolsep}{0.75pt}
{\small
\begin{tabular*}{\columnwidth}{@{\extracolsep{\fill}}lrr}\toprule
Statistic & Estimate & 95\% CI \\\midrule
Response Pearson & 0.619 & [0.492, 0.726] \\ 
Response Spearman & 0.572 & [0.458, 0.671] \\ 
Model Pearson & 0.817 & -- \\ 
Tier agreement & 80.0\% & -- \\ 
ICC(2,$k$) & 0.925 & -- \\ 
Within 25 points & 94.0\% & -- \\ 
\bottomrule\end{tabular*}}
\par\vspace{4pt}
{\small\begin{tabular*}{\columnwidth}{@{\extracolsep{\fill}}lrrrl}\toprule
Model & $n$ & Annotator & LLM & Tier (A/L) \\\midrule
\makecell[l]{Qwen3-Omni-\\Thinking} & 19 & 61.40 & 64.04 & Strong/Strong \\
Qwen3-Omni & 23 & 40.22 & 46.38 & Moderate/Moderate \\
OmniVinci & 25 & 35.67 & 35.67 & Moderate/Moderate \\
Qwen2.5-Omni & 25 & 32.33 & 34.00 & Moderate/Moderate \\
Gemini 3 Pro & 15 & 26.11 & 22.78 & Moderate/Poor \\
Gemini 2.5 Flash & 18 & 17.13 & 17.59 & Poor/Poor \\
Gemini 2.5 Pro & 19 & 16.23 & 16.67 & Poor/Poor \\
Gemini 3 Flash & 16 & 15.62 & 20.83 & Poor/Poor \\
VITA-1.5 & 29 & 10.92 & 41.67 & Poor/Moderate \\
Baichuan-Omni-1.5 & 11 & 10.61 & 18.94 & Poor/Poor \\
\bottomrule\end{tabular*}}

\end{table}

\FloatBarrier
\newcommand{\failurecaseblock}{%
\suppanchor{appx:failure_cases}
\begin{figure*}[t]
\begin{minipage}[t]{0.32\textwidth}
\small
\textbf{Who: speaker binding.} The target speaker is the man on the right during the marked interval. A salient close-up of the other participant can attract the prediction even when audio and later visual context identify the correct speaker.

\medskip
\centering
\includegraphics[width=\linewidth]{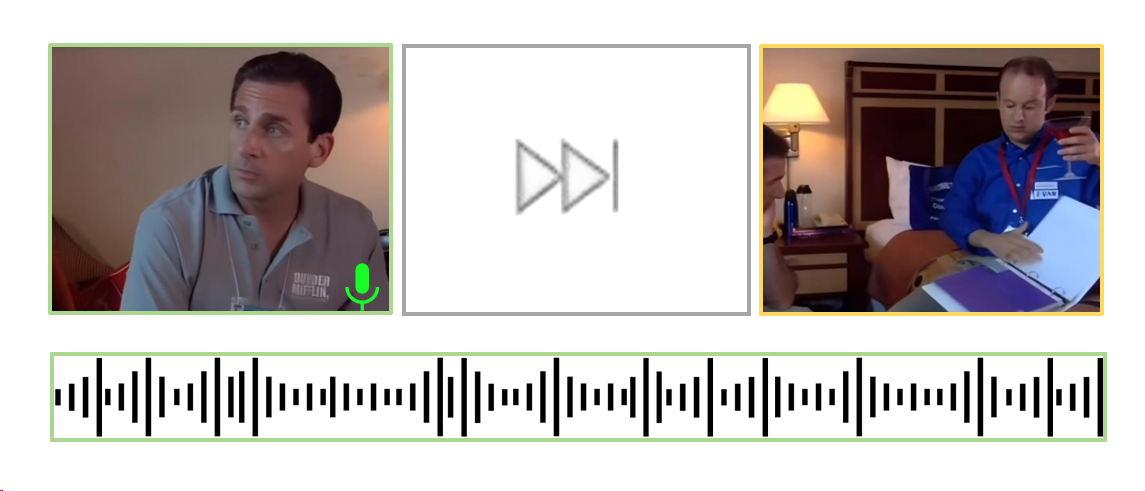}
\normalsize
\captionof{figure}{Representative \emph{who} failure. The visually dominant participant is not the active speaker.}
\label{fig:appx_failure_who}
\end{minipage}\hfill
\begin{minipage}[t]{0.32\textwidth}
\small
\textbf{When: premature entry.} The query lies in an ongoing turn containing a pause-like acoustic gap. The surrounding context remains pragmatically unfinished, but a fixed-prefix model can mistake the silence for turn completion.

\medskip
\centering
\includegraphics[width=\linewidth]{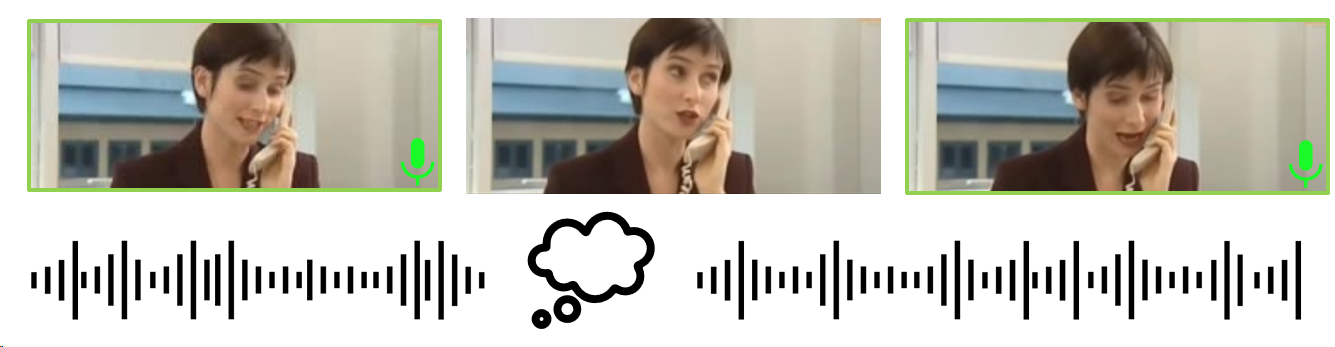}
\normalsize
\captionof{figure}{Representative \emph{when} failure. A pause-like gap creates a premature-entry trap.}
\label{fig:appx_failure_when}
\end{minipage}\hfill
\begin{minipage}[t]{0.32\textwidth}
\small
\textbf{How: pragmatic fit.} The target needs an empathetic continuation. Generic problem-solving advice can remain topically related while violating the target role and the social purpose of the exchange.

\medskip
\centering
\includegraphics[width=\linewidth]{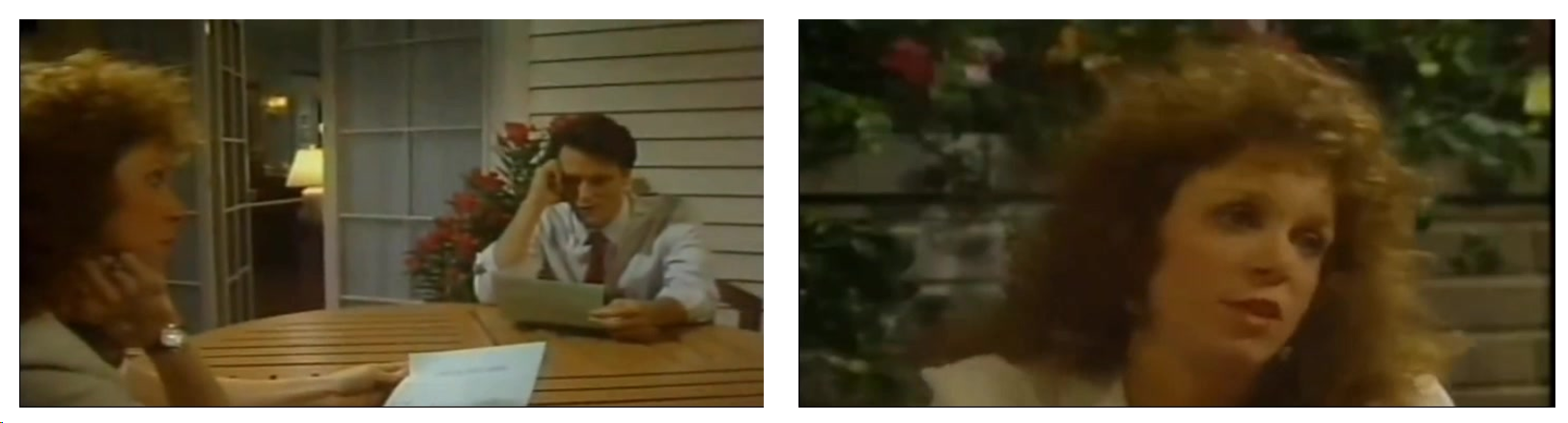}
\normalsize
\captionof{figure}{Representative \emph{how} failure. Topical relevance does not ensure role-consistent social fit.}
\label{fig:appx_failure_how}
\end{minipage}
\end{figure*}

\subsection{Representative Failure Cases}
\label{appx:failure_cases}

The three cases isolate complementary failures in cross-modal binding, continued-turn reasoning, and role-consistent generation. They are diagnostic examples rather than additional benchmark subsets.

Across the three cases, correct propositional content is insufficient: successful interaction also requires binding speech to the right participant, distinguishing hesitation from turn completion, and preserving the target role when generating a continuation.
\medskip
}
\failurecaseblock

\FloatBarrier
\clearpage
\suppanchor{appx:reliability_scope}
\section{Reliability, Bias, and Scope}
\label{appx:reliability_scope}

\suppanchor{appx:human_annotation_revision}
\subsection{Three-Annotator Verification}
\label{appx:human_annotation_revision}

Eight trained annotators participate in benchmark construction, with three independent checks assigned to every retained item before cross-review and senior adjudication. All task labels, dialogue text, and reference continuations are human-authored or manually verified. Raw pre-adjudication agreement is 94.2\% on perception and 91.8\% on interaction-generation annotations; Section~\ref{appx:iaa} details the annotation roles.

\suppanchor{appx:identity_consistency_revision}
\subsection{Target Identity and Visibility Labels}
\label{appx:identity_consistency_revision}

The target participant in \emph{when} is specified by a visible description rather than by the phrase ``your turn.'' For \emph{who}, speaker identity and spoken content are independently varied across the four options. Visibility labels distinguish a clearly attributable on-screen speaker from naturally occurring off-screen, reaction-shot, occluded, or weak-cue cases. These labels support error stratification but do not constitute synthetic conflict pairs.

\suppanchor{appx:judge_revision}
\subsection{Judge Alignment and Residual Bias}
\label{appx:judge_revision}

The scalable \emph{how} protocol requires complete GPT-4o, Gemini~2.5 Pro, and Qwen3-Omni scores for every eligible response. Pairwise agreement, leave-one-judge-out rankings, same-family removal, and the blind three-annotator calibration in Section~\ref{appx:human_feedback} support broad quality stratification, but do not establish human equivalence or eliminate related-family preference.

Judges see one human continuation to clarify role and intent, whereas evaluated models do not. The reference-free and held-out sanity controls are supplementary sensitivity analyses, not replacements for the fixed three-judge reference-aware scores in the main results. Reliability conclusions are therefore specific to that primary protocol and do not establish reference-free judging reliability. Continued blind three-annotator calibration remains necessary as judges and response distributions change.

\suppanchor{appx:scope_limitations}
\subsection{Benchmark Scope}
\label{appx:scope_limitations}

The 200 manually reviewed interaction-generation items support aggregate diagnosis rather than fine-grained per-domain ranking. Dense verification strengthens target identity, query-state, and response-quality labels, but does not remove uncertainty from set size, category coverage, or source diversity. Claims therefore remain limited to the evaluated clips, models, and protocol.

The offline fixed-prefix protocol does not measure persistent state, deployment latency, interruption recovery, or multi-turn adaptation. Its short, mostly third-person clips also leave multilingual and culture-specific norms, first-person interaction, backchannels, and longer dialogue for future evaluation. Interface gaps require similar care: the GPT-4o cascade preserves query-bounded transcript semantics but loses raw prosody, while the evaluated Gemini relay is visual-only. Native streaming and broader interaction settings are complementary extensions of this controlled cross-interface diagnostic.

\end{document}